\documentclass[10pt]{article}
\usepackage{latexsym,amsmath,amssymb,graphics,amscd, empheq}
\usepackage{enumitem}
\usepackage{mathbbol}
\usepackage{color}
\usepackage{stackengine}
\usepackage[breaklinks,colorlinks,
  citecolor=blue,
  urlcolor=blue]{hyperref}

\DeclareFontFamily{OT1}{pzc}{}
\DeclareFontShape{OT1}{pzc}{m}{it}{<-> s * [1.10] pzcmi7t}{}
\DeclareMathAlphabet{\mathpzc}{OT1}{pzc}{m}{it}

\addtocounter{footnote}{1}
\makeatletter
\@addtoreset{table}{bsection}
\def\thetable{\thesection.\@arabic\c@table}
\def\fps@table{h, t}
\@addtoreset{equation}{section}

\makeatother

\newtheorem{theorem}{Theorem}[section]

\providecommand{\norm}[1]{\lVert#1\rVert}

\newcommand{\vertiii}[1]{{\left\vert\kern-0.25ex\left\vert\kern-0.25ex\left\vert #1 
    \right\vert\kern-0.25ex\right\vert\kern-0.25ex\right\vert}}
\newsavebox{\savepar}

\makeatletter

\usepackage{scalerel,stackengine}
\stackMath
\newcommand\reallywidehat[1]{%
\savestack{\tmpbox}{\stretchto{%
  \scaleto{%
    \scalerel*[\widthof{\ensuremath{#1}}]{\kern-.6pt\bigwedge\kern-.6pt}%
    {\rule[-\textheight/2]{1ex}{\textheight}}
  }{\textheight}%
}{0.5ex}}%
\stackon[1pt]{#1}{\tmpbox}%
}

\begin{document}

\title{\textbf{Fading memory echo state networks are universal}}
\author{Lukas Gonon$^{1}$ and Juan-Pablo Ortega$^{2, 3}$}
\date{}
\maketitle

\begin{abstract}
Echo state networks (ESNs) have been recently proved to be universal approximants for input/output systems with respect to various $L ^p$-type criteria. When $1\leq p< \infty$, only $p$-integrability hypotheses need to be imposed, while in the case $p=\infty$ a uniform boundedness hypotheses on the inputs is required. This note shows that, in the last case, a universal family of ESNs can be constructed that contains exclusively elements that have the echo state and the fading memory properties. This conclusion could not be drawn with the results and methods available so far in the literature. 
\end{abstract}

\bigskip


\makeatletter
\addtocounter{footnote}{1} \footnotetext{%
Ludwig-Maximilians-Universit\"at M\"unchen. Mathematics Institute.
Theresienstrasse 39.
D-80333 Munich. Germany. {\texttt{gonon@math.lmu.de} }}
\addtocounter{footnote}{1} \footnotetext{%
Universit\"at Sankt Gallen. Faculty of Mathematics and Statistics. Bodanstrasse 6.
CH-9000 Sankt Gallen. Switzerland. {\texttt{Juan-Pablo.Ortega@unisg.ch}}}
\addtocounter{footnote}{1} \footnotetext{%
Centre National de la Recherche Scientifique (CNRS). France.}
\makeatother

\medskip

\medskip

\section{The problem}

\paragraph{Main goal.}
The objective of this note is showing that the universal families of 
echo state networks (ESNs) devised in the main theorem of \cite{RC7} can be chosen to satisfy the echo state and the fading memory properties. This fact was not established in the original paper and showing it requires a different strategy that will be presented in the proof of the main result later on in Section \ref{The result}.

\paragraph{Context and motivation.}
Reservoir computing \cite{lukosevicius, tanaka:review}  in general and ESNs~\cite{Matthews1993, Jaeger04} in particular have exhibited a remarkable success in the learning of the chaotic attractors of complex nonlinear infinite dimensional dynamical systems \cite{Jaeger04, pathak:chaos, Pathak:PRL, Ott2018, hart:ESNs, RC18} and in a great variety of empirical classification and forecasting applications \cite{Wyffels2010, Buteneers2013, GHLO2014}).  These findings have been an important motivation for the in-depth study of the approximation capabilities of these machine learning paradigms. The first results in this direction have  been obtained in the context of systems theory for input/output systems with either finite or approximately finite memory \cite{sandberg:esn, perryman:thesis, Stubberuda} in a forward-in-time framework. More recently, those universality statements were extended to ESNs with semi-infinite inputs from the past.  Various $L ^p$-type criteria have been used to measure the approximation error. The case $1\leq p< \infty$ has been considered in \cite{RC8}, where the universality of families of ESNs with a prescribed activation function and stochastic inputs was established with respect to the $L ^p$ norm determined by the law of a fixed discrete-time input process defined  for all infinite negative times. In this case, the universality is formulated in the category of all the causal and time-invariant input/output systems with $p $-integrable outputs. Extensions of some of these results in the particularly relevant case $p=2 $ for \textit{randomly} generated ESNs and, more importantly, corresponding approximation and generalization error bounds for regular filters/functionals have been derived in \cite{RC10, RC12}.

Universality with respect to uniform approximation, that is, the case $p= \infty $, has been studied in \cite{RC6, RC7} for (almost surely) uniformly bounded inputs in the fading memory category. The universality of ESNs for uniformly bounded inputs in the fading memory category has been established in \cite{RC7} using an internal approximation property  \cite[Theorem 3.1]{RC7} that allows one to conclude the uniform proximity of input/output systems generated by a state-space system out of the uniform closeness of the corresponding state maps. Using this observation, it can be shown that ESNs inherit universality properties out of the universality of neural networks \cite{cybenko, hornik}. 

The result that we just described, stated in Theorem 4.1 of \cite{RC7}, does not guarantee though that the approximating ESNs have two important properties that we now briefly recall. The first one is the echo state property (ESP) that holds when every semi-infinite input has one and only one semi-infinite output associated. The ESP allows, in passing, to associate a filter to the ESN. The second one is the fading memory property which, in the presence of uniformly bounded inputs, amounts to the continuity of the associated filter when the spaces of inputs and outputs are endowed with the product topologies. {\it The main theorem in this note shows that, when the activation function in the universal family of ESNs is Lipschitz-continuous, then this family can be chosen so that all its elements have the echo state and the fading memory properties.} This fact was not established in the original paper and showing it requires a different strategy than in \cite{RC7}.

\section{The result}
\label{The result}

The statement and proof of the main theorem uses a notation similar to that in \cite{RC7}. In particular, for any $M > 0 $, we denote by $B_{\left\|\cdot \right\|}({\bf 0}, M)$ the Euclidean ball of radius $M$  and by $ \overline{B_{\left\|\cdot \right\|}({\bf 0}, M)} $ its closure. The set $K _M:=\overline{B_{\left\|\cdot \right\|}({\bf 0}, M)} ^{\mathbb{Z}_{-}}$ is the Cartesian product of $\mathbb{Z}_{-} $ copies of $\overline{B_{\left\|\cdot \right\|}({\bf 0}, M)} $ endowed with the product topology. Given $M,L> 0 $ and $U: K_M\subset \left(\mathbb{R}^d\right)^{\mathbb{Z}_{-}} \longrightarrow K _L \subset  (\mathbb{R}^m)^{\mathbb{Z}_-}$ a causal and time-invariant filter, for some $d, m \in \mathbb{N} $, we  denote by $H _U:= p _0\circ U:K _M \longrightarrow \mathbb{R}^m $ the associated functional, with $p _0:\left(\mathbb{R}^m\right)^{\mathbb{Z}_{-}}\longrightarrow \mathbb{R}^m $ the projection onto the zero entry. The space of filters and functionals of the type of $U$  and $H _U $ can be endowed with a uniform norm $\vertiii{\cdot } _{\infty}$ (see (2.16) and (2.17) in \cite{RC7} for definitions) with respect to which we shall prove the universality of the echo state family with the properties described above.
We recall that an echo state network is determined by the state-space system:
\begin{equation}
\label{prescription esn}
\begin{cases}
{\bf x}_t = \boldsymbol{\sigma}(A{\bf x}_{t-1} + C {\bf z}_t + \boldsymbol{\zeta}),\\
{\bf y}_t = W {\bf x}_t,
\end{cases}
\end{equation}
where the states $\mathbf{x} _t \in \mathbb{R}^N $, for some $N \in \mathbb{N} $, $A \in \mathbb{M}_{N,N} $, $C \in  \mathbb{M}_{N,d} $, $\boldsymbol{\zeta} \in  \mathbb{R} ^N $, and $W \in \mathbb{M}_{m,N} $. The values ${\bf z} _t  \in \mathbb{R}^d $ (respectively, ${\bf y} _t \in \mathbb{R}^m$) are the components of the input sequence ${\bf z} \in K _M  $ (respectively, the output sequence ${\bf y} \in K _L  $). The map $\boldsymbol{\sigma}: \mathbb{R}^N \longrightarrow\mathbb{R}^N  $ is obtained by componentwise application of an activation function $\sigma: \mathbb{R}\longrightarrow\mathbb{R} $ that we assume $L_\sigma$-Lipschitz-continuous, bounded, and non-constant.

\begin{theorem}[Universality of the echo state family]
\index{Universality of the echo state family with invertible activation functions}
Let $M, L>0$, let $U: K_M \longrightarrow K _L$ be a causal and time invariant filter that has the fading memory property, and let  $\sigma: \mathbb{R}\longrightarrow\mathbb{R} $ be Lipschitz-continuous, non-constant, and bounded. Then, for any $\epsilon> 0  $ there exists an echo state network of the type \eqref{prescription esn}  with activation function determined by $\sigma$ that has the echo state and the fading memory properties and whose associated filter $U_{\text{ESN}}: K_M \longrightarrow (\mathbb{R}^m)^{\mathbb{Z}_-}$ satisfies that
\begin{equation*}
   \vertiii{U - U_{\text{ESN}}}_{\infty} < \epsilon.
\end{equation*}
\end{theorem}
\vspace{0.5cm}
\noindent{\bf Proof.\ }
Let $H_U: K_M \longrightarrow \mathbb{R}^m$ be the functional associated to the filter $U$ in the statement. There are various results in the literature (see, for instance, \cite[Theorems 3 and 4]{Boyd1985}, \cite[Remark 12]{RC6}, or \cite[Theorem 31]{RC9}) that guarantee that $H _U $ can be uniformly approximated by a finite memory functional. This means that for any $\epsilon > 0$ there exists $K\in \mathbb{N}$ and a continuous map $G: \overline{B_{\left\|\cdot \right\|}({\bf 0}, M)}^{K+1} \longrightarrow \mathbb{R}^m$ such that 
\begin{equation}
\label{eq: epsilon approximation in proof univ esn}
   \sup_{{\bf z}\in K_M} \left \{ \left\|H_U({\bf z})- G({\bf z}_{-K}, {\bf z}_{-K + 1}, \ldots, {\bf z}_0)\right\| \right \} < \frac{\epsilon}{3}.
\end{equation}
Let $D_d:= \overline{B_{\left\|\cdot \right\|}({\bf 0}, M)}\subset \mathbb{R} ^d$. As $G$ is continuous and it is defined on a compact set, the neural network approximation theorems (\cite[Theorem~2]{hornik1991}, see also \cite{cybenko, hornik}) imply the existence of a $\overline{N} \in \mathbb{N}, \overline{W} \in \mathbb{M}_{m, \overline{N}}, \mathcal{A} \in \mathbb{M}_{\overline{N}, (K+1)d}$, and $\overline{\boldsymbol{\zeta}} \in \mathbb{R}^{\overline{N}}$ such that, if we define $X:= \overline{B_{\left\|\cdot \right\|}({\bf 0}, M)}^{K+1} $, we have
\begin{equation}
\label{eq: approx of G by universal approximation}
   \sup_{{\bf u} \in X} \left \{ \norm {G({\bf u}) - \overline{W} \boldsymbol{\sigma}(\mathcal{A}{\bf u} + \overline{\boldsymbol{\zeta}})} \right \} < \frac{\epsilon}{3}.
\end{equation}
Let us partition the matrix as $\mathcal{A} =\left[A^{(-K)} A^{(-K+1)}\cdots A^{(0)}\right]$ with $A^{(-j)} \in \mathbb{M}_{\overline{N}, d}$ and define the constant $c = \norm{\overline{W}} L_{\sigma} \sum_{j=0}^K \norm{A^{(-j)}} j$. Again by the universal approximation theorem \cite{hornik1991} for each $j=1,\ldots,K$ there exists
$\widetilde{N}_j \in \mathbb{N}, \widetilde{W}_j \in \mathbb{M}_{d, \widetilde{N}_j}, \widetilde{A}_j \in \mathbb{M}_{\widetilde{N}_j,d}$, and $\widetilde{\boldsymbol{\zeta}}_j \in \mathbb{R}^{\widetilde{N}_j}$ such that the neural network $\mathcal{I}_j({\bf z}):= \widetilde{W}_j \boldsymbol{\sigma}(\widetilde{A}_j{\bf z} + \widetilde{\boldsymbol{\zeta}}_j)$ approximates the identity uniformly on $\overline{B_{\left\|\cdot \right\|}({\bf 0}, M+(j-1)\frac{\epsilon}{3c})}\subset \mathbb{R} ^d$, that is,  
\begin{equation}
\label{eq: approx of Identity by universal approximation}
\sup_{{\bf z} \in \overline{B_{\left\|\cdot \right\|}({\bf 0}, M+(j-1)\frac{\epsilon}{3c})}} \left \{ \norm {\mathcal{I}_j({\bf z}) -  {\bf z}} \right \} < \frac{\epsilon}{3c}.
\end{equation}
Define $\mathcal{J}_j = \mathcal{I}_j \circ \cdots \circ \mathcal{I}_1$ and $\mathcal{J}_0({\bf z}) = {\bf z}$. We now prove inductively that for each $j=1,\ldots,K$
\begin{equation} \label{eq: approx of Identity concatenation} \begin{aligned} 
\sup_{{\bf z} \in \overline{B_{\left\|\cdot \right\|}({\bf 0}, M)}} \left \{ \norm {\mathcal{J}_j({\bf z}) - {\bf z}} \right \} & < j \frac{\epsilon}{3c}.
\end{aligned} \end{equation}
For $j=1$ this follows from  \eqref{eq: approx of Identity by universal approximation}. For the induction step, we assume \eqref{eq: approx of Identity concatenation} holds for indices up to $j-1$ and aim to prove it for $j$. Firstly, we obtain from the induction hypothesis for $i=1,\ldots,j-1$
\[
\sup_{{\bf z} \in \overline{B_{\left\|\cdot \right\|}({\bf 0}, M)}} \left \{ \norm {\mathcal{J}_i({\bf z})} \right \}  \leq \sup_{{\bf z} \in \overline{B_{\left\|\cdot \right\|}({\bf 0}, M)}} \left \{ \norm {\mathcal{J}_i({\bf z}) - {\bf z}}\right \} + M \leq  i\frac{\epsilon}{3c} + M
\]
and hence $\mathcal{J}_{i}({\bf z}) \in \overline{B_{\left\|\cdot \right\|}({\bf 0}, M+i\frac{\epsilon}{3c})}\subset \mathbb{R} ^d$ for ${\bf z} \in D_d$. This inclusion, the triangle inequality, and \eqref{eq: approx of Identity by universal approximation} imply that
\begin{multline*}
\sup_{{\bf z} \in \overline{B_{\left\|\cdot \right\|}({\bf 0}, M)}} \left \{ \norm {\mathcal{J}_j({\bf z}) -  {\bf z}} \right \}   \leq 
\sum_{i=1}^j \sup_{{\bf z} \in \overline{B_{\left\|\cdot \right\|}({\bf 0}, M)}} \left \{ \left\| \mathcal{I}_i( \mathcal{J}_{i-1}({\bf z}))   - \mathcal{J}_{i-1}({\bf z}) \right\| \right \}\\  
\leq  \sum_{i=1}^j \sup_{{\bf z} \in \overline{B_{\left\|\cdot \right\|}({\bf 0}, M+(i-1)\frac{\epsilon}{3c})}} \left \{ \left\| \mathcal{I}_i( {\bf z})  - {\bf z} \right\| \right \}  < j \frac{\epsilon}{3c},
\end{multline*}
which proves \eqref{eq: approx of Identity concatenation}. The Lipschitz continuity of $\sigma$ and \eqref{eq: approx of Identity concatenation} thus allow us to estimate
\begin{multline}
\label{eq: inverse G in proof of theorem}
 \sup_{{\bf z}\in K_M} \left \{ \left\| \overline{W} \boldsymbol{\sigma}\left( \sum_{j=0}^K A^{(-j)} {\bf z}_{-j} + \overline{\boldsymbol{\zeta}}\right) -  \overline{W} \boldsymbol{\sigma}\left( \sum_{j=0}^K A^{(-j)} \mathcal{J}_j({\bf z}_{-j}) + \overline{\boldsymbol{\zeta}}\right) \right\| \right \} \\
\leq \norm{\overline{W}} L_{\sigma} \sum_{j=0}^K \norm{A^{(-j)}} \sup_{{\bf z}\in K_M} \left \{ \left\| {\bf z}_{-j} -  \mathcal{J}_j({\bf z}_{-j})    \right\| \right \} 
\leq \norm{\overline{W}} L_{\sigma} \sum_{j=0}^K \norm{A^{(-j)}} j \frac{\epsilon}{3c} = \frac{\epsilon}{3} .
\end{multline}
Let $H^{\text{ESN}}({\bf z}) := \overline{W} \boldsymbol{\sigma}\left(\sum_{j=0}^K A^{(-j)} \mathcal{J}_j({\bf z}_{-j}) + \overline{\boldsymbol{\zeta}}\right)$ and $H^{\text{FNN}}({\bf z}) := \overline{W} \boldsymbol{\sigma}\left(\sum_{j=0}^K A^{(-j)} {\bf z}_{-j} + \overline{\boldsymbol{\zeta}}\right)$. By the triangle inequality, \eqref{eq: epsilon approximation in proof univ esn}, \eqref{eq: approx of G by universal approximation}, and \eqref{eq: inverse G in proof of theorem}, we have 
\begin{equation*}\begin{aligned}
   \vertiii{H_U-H^{\text{ESN}}}_\infty & =\sup_{{\bf z} \in K_M} \{ \left\|H_U({\bf z})-H^{\text{ESN}}({\bf z})\right\| \} \\ & \leq 
   \vertiii{H_U-G}_\infty + \vertiii{G-H^{\text{FNN}}}_\infty + \vertiii{H^{\text{FNN}}-H^{\text{ESN}}}_\infty < \frac{\epsilon}{3}+ \frac{\epsilon}{3} + \frac{\epsilon}{3}= \epsilon.
   \end{aligned}
\end{equation*}
In order to conclude the proof, it remains to be shown that $H^{\text{ESN}}$ is indeed the functional associated to an echo state network. Let  $N=\widetilde{N}_1 \cdots + \widetilde{N}_K +\overline{N} $ and
\medskip
\begin{equation*}
\footnotesize
 \!\!\! \!\!\! \!\!\! \!\!\! \!\!\! \!\!\!A = \left(
\begin{array}{cccccc}
\mathbb{0}_{\widetilde{N}_1,\widetilde{N}_1}&\mathbb{0}_{\widetilde{N}_1,\widetilde{N}_2} & \cdots &\cdots & \mathbb{0}_{\widetilde{N}_1,\widetilde{N}_K} & \mathbb{0}_{\widetilde{N}_1,\overline{N}}\\
	& & & & &\\
\widetilde{A}_2 \widetilde{W}_1 & \mathbb{0}_{\widetilde{N}_2,\widetilde{N}_2} & \cdots& \cdots & \mathbb{0}_{\widetilde{N}_2,\widetilde{N}_K} & \mathbb{0}_{\widetilde{N}_2,\overline{N}}
\\
\mathbb{0}_{\widetilde{N}_3,\widetilde{N}_1} & \widetilde{A}_3 \widetilde{W}_2& \ddots & \ddots& \mathbb{0}_{\widetilde{N}_3,\widetilde{N}_K} & \mathbb{0}_{\widetilde{N}_3,\overline{N}} 
\\
\vspace{-0pt}	& & & & & \\
\vdots & \vdots & \ddots & \ddots & \vdots& \vdots \\

\vspace{-0pt}	& & & & & \\
\mathbb{0}_{\widetilde{N}_K,\widetilde{N}_1} & \mathbb{0}_{\widetilde{N}_K,\widetilde{N}_2}  & \cdots &\widetilde{A}_K \widetilde{W}_{K-1} & \mathbb{0}_{\widetilde{N}_K,\widetilde{N}_K} & \mathbb{0}_{\widetilde{N}_K,\overline{N}}\\
	& & & & & \\
A^{(-1)}\widetilde{W}_{1}& A^{(-2)} \widetilde{W}_{2} &  \cdots & A^{-(K-1)} \widetilde{W}_{K-1} & A^{(-K)} \widetilde{W}_{K} & \mathbb{0}_{\overline{N},\overline{N}}
\end{array}
\right) \in \mathbb{M}_{N, N},\ 
C = \left(
\begin{array}{c}
\widetilde{A}_1\\
\mathbb{0}_{\widetilde{N}_2,d}\\
\vdots\\
\mathbb{0}_{\widetilde{N}_K,d}\\
\vspace{-5pt}	\\
A^{(0)}
\end{array}
\right) \in \mathbb{M}_{N, d},\ 
\boldsymbol{\zeta} = \left(
\begin{array}{c}
\widetilde{\boldsymbol{\zeta}}_1\\
\widetilde{\boldsymbol{\zeta}}_2\\
\vdots\\
\widetilde{\boldsymbol{\zeta}}_K\\
\vspace{-6pt}	  \\
\overline{\boldsymbol{\zeta}}
\end{array}
\right) \in \mathbb{R}^{N}.
\end{equation*}
Consider now the state vectors 
\begin{equation*}
{\bf x}_t = \left(
\begin{array}{c}
\mathbf{x}_t^{(0)}\\
\mathbf{x}_t^{(1)}\\
\vdots\\
\mathbf{x}_t^{(K)}
\end{array}
\right) \in \mathbb{R}^N\  \mbox{with $\mathbf{x}_t^{(i)} \in \mathbb{R}^{\widetilde{N}_{i+1}} $, $i \in \{0, 1, \ldots, K-1 \} $, $\mathbf{x}_t^{(K)} \in \mathbb{R}^{\overline{N}}  $,}
\end{equation*}
and the state equation in $\mathbb{R}^N $ determined by
\begin{equation}\label{eq: state eq in proof univ esn}
   {\bf x}_t = \boldsymbol{\sigma}(A{\bf x}_{t-1} + C{\bf z}_t + \boldsymbol{\zeta}).
\end{equation}
By our choice of matrices, the solutions ${\bf x}$ of \eqref{eq: state eq in proof univ esn} (if they exist) satisfy that
\begin{multline}
\label{eq: iterating eq}
   {\bf x}_t^{(0)} = \boldsymbol{\sigma}(\widetilde{A}_1 {\bf z}_t + \widetilde{\boldsymbol{\zeta}}_1) \in \mathbb{R}^{\widetilde{N}_1},\, {\bf x}^{(j)}_t = \boldsymbol{\sigma}\left(\widetilde{A}_{j+1} \widetilde{W}_{j}{\bf x}_{t-1}^{(j-1)} + \widetilde{\boldsymbol{\zeta}}_{j+1} \right) \in \mathbb{R}^{\widetilde{N}_{j+1}}, \  \mbox{with} \  j\in \{1, \ldots, K-1\}, \  \mbox{and}\\
   {\bf x}^{(K)}_t = \boldsymbol{\sigma} \left(\sum_{j=1}^K A^{(-j)} \widetilde{W}_{j} {\bf x}_{t-1}^{(j-1)} + A^{(0)}{\bf z}_t + \overline{\boldsymbol{\zeta}} \right) \in \mathbb{R}^{\overline{N}}, \  \mbox{for all} \  t \in \mathbb{Z}_{-}.
\end{multline}
Iterating the expression \eqref{eq: iterating eq} we obtain that this solution indeed exists, it is unique, and it is given by:
\begin{multline*}
   {\bf x}_t^{(0)} = \boldsymbol{\sigma}(\widetilde{A}_1 {\bf z}_t + \widetilde{\boldsymbol{\zeta}}_1),\, {\bf x}^{(1)}_t = \boldsymbol{\sigma}(\widetilde{A}_2 \mathcal{I}_1({\bf z}_{t-1})+\widetilde{\boldsymbol{\zeta}}_2 ),\, \ldots, {\bf x}^{(K-1)}_t = \boldsymbol{\sigma}(\widetilde{A}_K \mathcal{J}_{K-1}({\bf z}_{t-(K-1)}) + \widetilde{\boldsymbol{\zeta}}_K), \quad \mbox{and} \\
{\bf x}_t^{(K)} =  \boldsymbol{\sigma}\left(\sum_{j=1}^{K} A^{(-j)} \widetilde{W}_j \boldsymbol{\sigma}\left(\widetilde{A}_{j} \mathcal{J}_{j-1}({\bf z}_{t-j}) + \widetilde{\boldsymbol{\zeta}}_{j} \right) + A^{(0)}{\bf z}_t+ \overline{\boldsymbol{\zeta}}\right)
=  \boldsymbol{\sigma}\left(\sum_{j=0}^{K} A^{(-j)} \mathcal{J}_j({\bf z}_{t-j}) + \overline{\boldsymbol{\zeta}}\right).
\end{multline*}
Consequently, 
$$ \overline{W}{\bf x}_0^{(K)} = \overline{W} \boldsymbol{\sigma}\left ( \sum_{j=0}^K A^{(-j)} \mathcal{J}_j ({\bf z}_{-j}) + \overline{\boldsymbol{\zeta}}\right) = H^{\text{ESN}}({\bf z}).$$
This equality shows that  $H^{\text{ESN}}$ is the functional associated to the echo state equation \eqref{eq: state eq in proof univ esn} and the linear readout given by $W := \left(\mathbb{O}_{m, N-\overline{N}}| \overline{W}\right) \in \mathbb{M}_{m, N}$, which concludes the proof.
\quad $\blacksquare$

\bigskip

\noindent {\bf Acknowledgments:} The authors acknowledge partial financial support  coming from the Research Commission of the Universit\"at Sankt Gallen, the Swiss National Science Foundation (grant number 200021\_175801/1), and the French ANR ``BIPHOPROC" project (ANR-14-OHRI-0002-02). The authors thank the hospitality and the generosity of the Division of Mathematical Sciences of the Nanyang Technological University, Singapore, and the FIM at ETH Zurich, where a significant portion of the results in this paper was obtained.

\bibliographystyle{wmaainf}


\end{document}